\documentclass[11pt,a4paper]{article}

\usepackage[nohyperref]{acl2020}
\usepackage{times}
\usepackage{latexsym}
\usepackage{amsmath}
\usepackage{amsfonts}
\usepackage[bottom]{footmisc}
\usepackage{graphicx}

\usepackage{xcolor}
\definecolor{medium-blue}{rgb}{0,0,1}
\usepackage{multirow}
\usepackage{adjustbox}

\usepackage{microtype}

\aclfinalcopy 

\title{Stock Index Prediction with Multi-task Learning and Word Polarity Over Time}

\author{Yue Zhou\qquad
  Kerstin Voigt\\
  School of Computer Science and Engineering, \\
  California State University, San Bernardino, CA, USA  \\
  \texttt{\{Yue.Zhou,Kvoigt\}@csusb.edu} \\
}

\date{}

\begin{document}
\maketitle
\begin{abstract}
  Sentiment-based stock prediction systems aim to explore sentiment or event signals from online corpora and attempt to relate the signals to stock price variations. Both the feature-based and neural-networks-based approaches have delivered promising results. However, the frequently minor fluctuations of the stock prices restrict learning the sentiment of text from price patterns, and learning market sentiment from text can be biased if the text is irrelevant to the underlying market. In addition, when using discrete word features, the polarity of a certain term can change over time according to different events. To address these issues, we propose a two-stage system that consists of a sentiment extractor to extract the opinion on the market trend and a summarizer that predicts the direction of the index movement of following week given the opinions of the news over the current week. We adopt BERT with multitask learning which additionally predicts the worthiness of the news and propose a metric called Polarity-Over-Time to extract the word polarity among different event periods. A Weekly-Monday prediction framework and a new dataset, the 10-year Reuters financial news dataset, are also proposed. 
  
\end{abstract}

\section{Introduction}
Over the past decades, sentiment analysis has delivered promising results on stock market prediction. The sentiment-based approaches explore the sentiment or event signals from text corpus such as tweets, news, or financial announcements and attempt to relate such signals to the stock price variation. These works can be regression~\citep{schumaker2009textual,bollen2011twitter} that predicts the \emph{price value} in next time step, or classification~\citep{xie2013semantic,smailovic2013predictive,ding2015deep,hu2018listening} that predicts the \emph{direction} of the stock movement. 

In early studies, bag-of-words (BoW), n-grams, or other discrete word features such as noun phrase are used to represent the text corpus~\citep{antweiler2004all,tetlock2008more,bollen2011twitter}. The word features are selected by pre-defined dictionaries or statistical metrics. Although, such approaches facilitate the alignment between the linguistic features and numerical data and mitigate the dimensionality problem, they can hardly preserve the contextual information.

Lately, neural approaches have been applied to the realm. \citet{ding2015deep} extracted event tuples (who did what to whom) from the news articles and trained the event embeddings. Then the daily events are averaged with the event vectors. \citet{pagolu2016sentiment} represented each tweet using Continuous Bag of Words Model (CBOW) with word2vec for prediction. Moreover, many neural network based models such as RNN and text-CNN are proposed~\citep{ding2015deep, vargas2017deep,hu2018listening,xu2018stock} and achieve considerable improvement as compared to traditional methods.

However, some issues are less addressed by existing approaches: (1) Most of the time the stock prices fluctuate within a narrow range and unnecessarily reflect the market sentiment, which complicates relating sentiment signals to price signals. (2) The polarity of a certain word, especially an entity or a term, can change over time according to different events. For instance, the tone of word ``venezuela" fluctuates with the oil trade situations between the U.S. and Venezuela. (3) The quality or value varies in different texts. Extracting sentiment signals from a text can only be valid if the text is relevant to the market. 

\begin{figure}[!htb]
    \centering 
    \includegraphics[width=\columnwidth]{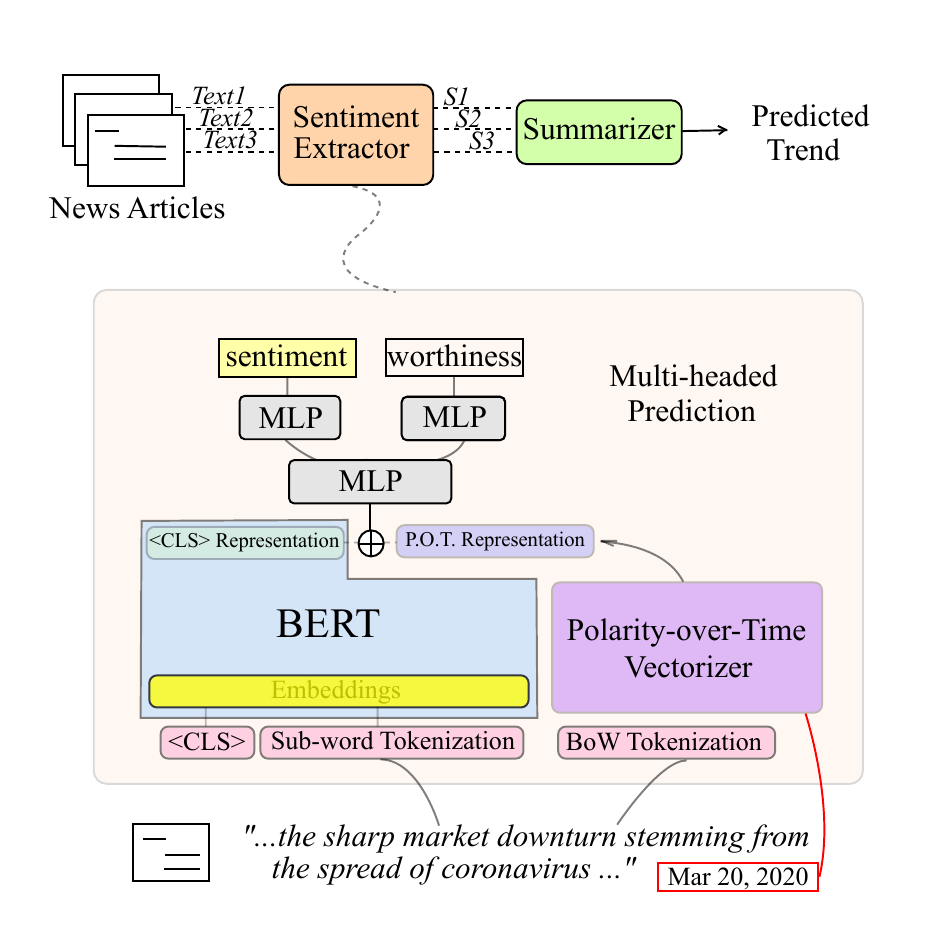}
    \caption{An overview of our prediction system.}
    \label{fig:flowc}
\end{figure}

To address these issues, we propose a two-stage system and utilize both the neural representation and discrete features for stock trend prediction. As illustrated in Figure~\ref{fig:flowc}, our system consists of a \emph{Sentiment Extractor} to extract the sentiment score for the future market trend and a \emph{Summarizer} that predicts the direction of the index movement of next week given the sentiment scores of news over the week. The main architecture we use for the sentiment extractor is the vanilla BERT~\citep{devlin2018bert} with multitask learning~\citep{caruana1997multitask,ruder2017overview} - one additional prediction head that predicts the worthiness of the news. We propose a metric called \emph{Polarity-Over-Time} to extract the word polarity among different event periods and use it as a supplemental feature. We present a weekly-Monday prediction framework in which the Monday index variations are predicted with all news articles over the past week. A new dataset, the 10-year Reuters financial News (2009-2020), is also proposed. 

The experimental results on the 10-year Reuters financial news dataset show that our system achieves significant improvement as compared to the baseline
methods. We illustrate the weekly-Monday basis is appropriate for sentiment-based stock prediction. We show that our model that uses word polarity features and additionally learns the worthiness of the news can better predict the stock market index.

\section{Related Work}

Many sentiment-based stock market prediction systems have been proposed
and achieved promising results. They can be regressions that combine the extracted linguistic features with the numerical data and predict the price value of next time step, or classifications in which the text data are first labeled with the price trends (\emph{e.g.}, up or down, positive or negative), then the model is trained to predict the direction of the price movement given the texts. These approaches also vary in the time interval for prediction, the corpus to use, and more importantly, the processing and representation of the corpus. Table~\ref{tab:related-work-list} shows an example of some representative works and their variations.

\begin{table*}[!htb]
\centering
\begin{adjustbox}{width=\textwidth}
\begin{tabular}{lllll}
\hline
\textbf{Work} & \textbf{Interval} & \textbf{Task} & \textbf{Corpus} & \textbf{Treatment} \\
\hline

\citet{schumaker2009textual} & Intraday & Regression & News & noun phrases  \\

\citet{bollen2011twitter} & Daily & Regression & Tweets & n-gram/lexicon  \\

\citet{xie2013semantic} & Daily & Classification & News & Frame/BoW/PoS  \\

\citet{ding2015deep} & Daily+Long/Short Term & Classification & News & EventsOfSeq \\

\citet{pagolu2016sentiment} & Daily + 3-day lag & Classification & Tweets & CBOW  \\

\citet{hu2018listening} & Daily + 10-day lag & Classification & News & CBOWOfSeq  \\

\citet{xu2018stock} & Daily + 5-day lag & Classification & Tweets & SeqOfSeq \\
\hline

\end{tabular}
\end{adjustbox}
\caption{Some sentiment-based models for stock prediction. Different approaches can vary in prediction interval, task definition, corpus, the process and representation of the corpus, and the model architecture.
}
\label{tab:related-work-list}
\end{table*}

In early studies, bag-of-words (BoW) based models \citep{antweiler2004all,mittermayer2004forecasting,das2007yahoo,tetlock2008more} were proposed. In these works, the word features are selected by pre-defined dictionaries or statistical metrics. However, n-grams and noun phrases can also be useful to capture the sentiment or event signals. \citet{schumaker2009textual} extracted noun phrases from financial news releases as additional features for the regression model that predicts the stock price for the next 20 minutes. \citet{bollen2011twitter} used tweets as corpus and computed six mood scores (Calm, Alert, Sure, Vital, Kind, and Happy) based on n-grams and lexicons, then use the scores to predict the index change of the next day. \citet{xie2013semantic} predicted the directional changes of the price based on financial news. In their work, multiple linguistic features (Semantic frames, bag-of-Words, part-of-speech DAL score) are extracted to capture particular events such as what happened to which company. To some extent, considering corpus as discrete features facilitates the alignment between the linguistic features and historical market data and solves the dimensionality problem (\emph{e.g.}, mapping millions of tweets per day to the price change of next day). However, such approach can hardly preserve the contextual information.

With the development of neural approaches for NLP, many works look for neural representations of the text and neural networks as model architectures. \citet{ding2015deep} extracted event tuples from the news articles by dependency parsing and pre-trained the event embeddings. Then the daily events are averaged with the event vectors and fed to the a CNN. \citet{pagolu2016sentiment} represented each tweet using Continuous Bag of Words Model (CBOW) with word2vec. \citet{hu2018listening} proposed a hybrid attention RNN for stock trend prediction based on the sequence of recent related news. Each news article is embedded as CBOW with word2vec. The RNN consisted of a time level attention (which time periods are more critical) and a context level attention (which news is more significant). Based on their work, \citet{xu2018stock} proposed a RNN based model, the stockNet, which takes the embedded stock messages and price differences as input. Each message is extracted from the preceding and following contexts of a stock symbol and represented by a bi-GRU. Then all messages in a day construct the event matrix. Finally, some creative variations are proposed. \citet{nguyen2015sentiment} used message board for certain stocks as the corpus rather than news or tweets; \citet{makrehchi2013stock} aggregated the predicted polarity of all tweets in a day as the stock movement signal. 

Despite their success, there are some issues to be addressed. When labelling the corpus with price trends, the data are imbalanced and extracting sentiment signals from minor fluctuations is difficult. The imbalanced labelling also leads to the difficulty of measuring the model performance. Moreover, the polarity of an entity or a term can change over time according to different events. Finally, some approaches ignore the heterogeneity of news articles or tweets - some text actually discuss the market while others are irrelevant. To address these issues, we propose a sentiment extractor that utilizes word polarity over time and multitask learning and a summarizer for stock index prediction.

\section{Methodology}

In this section, we will introduce the collection and labelling of the dataset we use and our model in details. We collected approximately 180,000 financial news articles from Reuters over the past ten years with respect to different companies, regions, and sectors. We define our goal as classification that predicts the S\&P 500 index trend of next week given the collection of news articles over the past week. Thus, the news data are labelled with the weekly difference of the S\&P 500 close prices. Our system will first learn to extract the sentiment of a news passage towards the stock market, then collectively use the overall sentiment information of the weekly news to predict the stock trend. 

\subsection{Dataset}

Our Reuters news collection contains 181,523 news articles from August 2009 to May 2020 and covers the following categories: 

\begin{itemize}
\item \textbf{Regional Market News}: United States, Europe and Middle East, and Asia Pacific

\item \textbf{Company News}: Companies in the Dow Jones Industrial Average and Top 50 Companies in the S\&P 500
\item \textbf{Sector News}: Financials, Technology, Industrials, Energy, Healthcare, Telecoms, Utilities, Basic Materials, Cyclical Goods \& Services, and Non-cyclical Goods \& Services.
\end{itemize}

\noindent Some news entries may belong to multiple categories (\emph{e.g.}, a news about Apple may also be found in the technology sector) or be edited and republished. Each entry at least has the following attributes: url of the news, title, content, published datetime, flags of company, region, or sector. The dataset can be obtained from \href{https://figshare.com/s/b74e77c172ddc78d2aef}{https://figshare.com/s/b74e77c172ddc78d2aef}.

\subsection{A Weekly-Monday Framework}

We use a weekly prediction interval and adopt a Monday-to-Monday basis. Concretely, our model predicts whether the close price of the S\&P 500 on Monday will increase or decrease as compared to last Monday. Given the Monday-to-Monday index difference, the news articles in a certain week will be grouped into positive, negative, or neutral for classification. 

We adopt such prediction framework for two reasons. First, a weekly interval better explores the market sentiments as a driving force in the short term. The investors need time to digest and then react on the news releases. Or, there is a latency for the news sentiments to ``contaminate" the market behavior. Second, the abnormal behavior of the Monday price or return has been analyzed by many works~\citep{cross1973behavior,jaffe1989twist}. The overall attitude of the market is gradually condensed as the weekend approaches when the market is closed and the investors start to prepare for the trading decisions on the upcoming Monday. 

Such hypothesis can be justified by two observations. First, the Monday-to-Monday prices are least tractable among all other weekdays. Table~\ref{table:m2mAClag} shows the autocorrelations of the daily close prices given five weekdays and different lags. As lag periods increases, the autocorrelations become weaker, however, the autocorrelation of the Monday price series always has the \emph{lowest} value. The index on Monday is most distorted by the uncertainty for which more space is left for sentiment analysis.

\begin{table}[!htb]
\centering

\begin{adjustbox}{width=\columnwidth}
\begin{tabular}{lccccc}
\hline
Lags & \textbf{Mon.}  & \textbf{Tue.} & \textbf{Wed.} & \textbf{Thu.} & \textbf{Fr.}\\
\hline
1 & 0.995197  & 0.995793 & 0.996149 & 0.995967 & 0.995989\\
5 & 0.977303   & 0.98005  & 0.980088  & 0.979865  & 0.979683 \\
10 &  0.961518  & 0.964764  & 0.965326  & 0.96493  & 0.963748 \\
20 &  0.895662  &  0.903705 & 0.899852  &  0.899938 & 0.898441 \\
40 &  0.772131  & 0.790959  & 0.790128  &  0.785565 & 0.782916 \\
\hline 
\end{tabular}

\end{adjustbox}
\caption{Autocorrelation of the price series on different weekdays.}
\label{table:m2mAClag}
\end{table}

Second, after grouping the news in positive and negative class according to the Monday price variations, we can look at the some representative uni-grams for the two classes by simply using the TF-IDF difference. As shown in Table~\ref{table:m2mwords}, besides the words illustrating market confidence such as ``gain" and ``loss", the words representing certain entities or events express strong sentiment. For example, ``coronavirus" and ``outbreak" indicate the market fear of 
the uncertain future, while ``lockdown", ``pandemic", and ``ventilator" imply the 
rebuild of confidence from the acknowledgement to the epidemic. This intrinsic evaluation justifies that the weekly interval especially the Monday-to-Monday basis is appropriated for sentiment-based stock prediction.

\begin{table*}
\centering
\small
\begin{tabular}{|l|l|l|}
\hline
  & \textbf{Positive} & \textbf{Negative} \\
\hline

Event/ & lockdown(s), pandemic, ventilator,  & coronavirus, outbreak, virus,  \\
Entity & distancing, raytheon, google, apple, & tariff(s), china, shanghai, trump,  \\
~ & tesla, mexico, etc.  & huawei, qualcomm, shell, gas, \\
~ & ~  & anadarko, exxon, khashoggi, iran,\\
~ & ~  & maduro, venezuela, etc.\\
\hline
Confidence/ & rise, gain(s), high(er), jump(ed), & fell, low(est), cut, impact, global\\
Periodicity & recovery, earnings, advance(d),  &  fears, lost, concern(s), drop(ed),\\
~ & export, jobs, employment, etc. & loss(es), etc.  \\
~ &  optimism, quarter, etc. &    \\


\hline

\end{tabular}
\caption{
Examples of positive and negative words selected by TF-IDF differences on the Monday-to-Monday basis. An entity often implies certain related events that influence the market. Raytheon was a major U.S. defense contractor and industrial corporation with core manufacturing concentrations in military; anadarko refers to the Anadarko Petroleum Corporation; Qualcomm is an American corporation manufacturing semiconductors, which is involved in the rival against Huawei. }
\label{table:m2mwords}
\end{table*}

\subsection{Model}

Our model consists of two parts: A sentiment extractor to extract the sentiment of a news article towards the upcoming/recent stock market and a summarizer that predicts the stock trend of next week given the sentiment of all news over the week.

\subsubsection{Sentiment Extractor}

The sentiment extractor aims to extract the sentiment scores towards the market from the news text. It is trained as binary classification on the positive and negative news that are labelled according to the approach discussed in section 3.2. We use the vanilla transformer~\citep{vaswani2017attention} as our main model architecture. Additional word features and multi-task learning are used to improve the model performance.

\paragraph{Word Polarity Over Time:}As mentioned in section 3.2, discrete word features can play a vital roll in sentiment detection. Although the words representing market confidence may always have consistent polarity (\emph{e.g.}, rise, fell, optimism, concern, etc.), the tone of the named entities changes over time according to the financial events it related to. For example, the word ``huawei" can be positive when the ban policy 
is in force, or negative when the countermeasures of Huawei take effect afterwards. The sentiment of word varies according to certain events at different periods of time. 

Thus, we introduce the Polarity-Over-Time (POT) to keep track of the polarity of a word in each week. For each week $t$, we group all the news articles over the past three months into five classes according to the weekly index changes: the very positive, positive, neutral, negative, and very negative. Then, the POT of word $x$ at week $t$, or $P_x^t$ is defined as:

\begin{equation} \label{pot}
\begin{split}
P_x^t = \frac{W_{x,vpos}^{t}}{\sqrt{N_{vpos}^{t}}}-\frac{W_{x,vneg}^{t}}{\sqrt{N_{vneg}^{t}}} \\ +  \alpha(\frac{W_{x,pos}^{t}}{\sqrt{N_{pos}^{t}}}-\frac{W_{x,neg}^{t}}{\sqrt{N_{neg}^{t}}} )
\end{split}
\end{equation}
\noindent where $W_{x,d}$ and $N_d$ ( $d = \{vpos, vneg, pos, neg\}$ ) are the TF-IDF of the word $x$ in document $d$ and the total number of document $d$, respectively. $\alpha$ is the discount factor. A greater value indicates a more positive polarity of the word and vise versa, and the polarity can change since the corpus changes on a quarterly rotation for each week. Finally, a news article can be represented by a $V\times L$ matrix, $M$ as:
\[
M = \begin{bmatrix}
P^t_{x_1} & P^{t-1}_{x_1} & \cdots  &P^{t-L+1}_{x_1} \\ 
P^t_{x_2} & P^{t-1}_{x_2} & \cdots  &P^{t-L+1}_{x_2} \\ 
 \vdots &  \vdots  & \ddots  &  \vdots \\ 
P^t_{x_v} & P^{t-1}_{x_v} & \cdots  &P^{t-L+1}_{x_v} \\ 
\end{bmatrix}
\]

\noindent where $V$ is the vocabulary size and $L$ is the number of lags. That is, the first column contains the POT scores of each word at week $t$ in which the news was published, and the second column contains the POT scores one week before, and so on. Such representation not only contains the polarity of words over time but also implicitly embeds the price movement information of previous weeks. Then we apply 
an attention function to get the vector representation of POT:

\begin{equation} \label{softmax}
a=so\hspace{-0.1em}f\hspace{-0.1em}tmax(v^T\!\tanh (W\!M))
\end{equation}
\begin{equation} \label{projectlinear}
V_{pot}= Ma^T
\end{equation}
\noindent where $W\in \mathbb{R}^{V\times V}$, $v\in \mathbb{R}^{V\times 1}$ are the weight parameters and $V_{pot}\in \mathbb{R}^{V\times 1}$. Finally, the full representation of a news is defined as: 

\begin{equation} \label{fullv}
V_{news} = V_{cls} \oplus V_{pot}
\end{equation}

\noindent where $V_{cls}$ is the pooled output (i.e., the representation of the CLS token for classification) of BERT.

\paragraph{Multitask Learning:}

To better learn to extract the market sentiment from a news report, it is useful to know if the news is actually relevant and important to the underlying market. We apply a multi-headed model to predict not only the sentiment but also the worthiness of the news. The worthiness prediction is a binary classification. Some labels are obtained by combining some of the existing categorical labels discussed in section 3.1. For example, a news from the basic-materials category: \emph{``Monsanto said it was cooperating with ongoing inquiries from the U.S. Justice Department about allegations it illegally dominates the market for genetically modified seeds..."} can be less relevant to the S\&P 500 index, as Monsanto is a low-volumne stock and replaced by Twitter in S\&P 500 in 2018, thus it is a robust proxy of a negative example of worthiness. In contrast, the news in terms of top companies in S\&P 500 or DIJA 30, technology and financial sectors, or the U.S region market can be used as positive examples of the worthiness. However, we also manually labelled some irrelevant news articles from these most seemingly important categories as negative examples to improve the quality of labelling. For example, a news article about Apple: \emph{``How about a beer with your iPhone?..."} is irrelevant to the stock market because it actually discussed a iphone case production that is equipped with a bottle opener.   

For worthiness, only a fraction of data are labelled due to the labor constraint. That is, a proportion of data for training the sentiment extractor have a pair of labels (sentiment, worthiness), while for the rest entries, the label of worthiness is unknown. We adopt multi-headed model and use cross entropy loss for both the sentiment head and worthiness head. Therefore, the loss function became a weighted sum of the two cross entropy losses:

\begin{equation} \label{loss} Loss(\theta) = \lambda C\hspace{-0.1em}E_{senti} + (1-\lambda) C\hspace{-0.1em}E_{worth}
\end{equation}

\noindent where $\theta$ denotes the parameters of the model, $CE_{senti}$ and $CE_{worth}$ are the cross entropy loss of the sentiment and worthiness classification, respectively. $\lambda$ is the weighting parameter. For the data whose worthiness label is unknown, the worthiness loss is ignored.

To conclude, as illustrated in Figure, the sentiment extractor take advantages of both the BERT representation and the Polarity-over-Time representation, and predicts the sentiment and worthiness of the news at the same time. We utilize such combined representation and multi-task learning so as to improve the model performance.

\subsubsection{Summarizer}

After the sentiment extractor is trained, we use the output from the softmax layer as the sentiment score for each news. For a certain week, we randomly select $N$ news articles and average their sentiment scores as the overall sentiment of the week. Thus, the goal of our summarizer is to learn a function that maps the overall sentiment of a week onto the index trend for next week. A simple SVM is used as the classifier. In the process of preparing data for training the summarizer, we only run the sentiment extractor on the news articles in the unseen weeks. This is because if the overall sentiment is collected from a week in which some news have been used to train the sentiment extractor, the price trend information will be leaked to the summarizer.  

Although the sentiment extractor is trained on the news published in the weeks experiencing greater price variations, it is reasonable to run the extractor on the news in ``neutral" weeks. This is because: (1) The news articles in the ``neutral" weeks does not necessarily have no polarity, rather, they are just the news articles published in the weeks when weekly index variation is small; (2) The sentiment extractor trained on the news weeks when big events happened can better learn to identify the event and sentiment signals. On the other hand, if the sentiment extractor is trained on both positive and negative news as well as the neutral news, a highly imbalanced classification problem occurs and the sentiment of individual news can hardly be learned from weak price signals.

\section{Experiments}

\subsection{Datasets and Evaluation Metrics}

We evaluate our model on the 10-year Reuters News dataset and the S\&P 500 index close price\footnote{\href{https://finance.yahoo.com/quote/\%5EGSPC/}{https://finance.yahoo.com/quote/\%5EGSPC/}} from August 2009 to May 2020. 161,989 out of 181,523 news articles are selected with bad data removed (advertisements, videos, long stories, etc.). We leave the data from May 2019 to May 2020 untouched for further analysis and split the data from August 2009 to April 2019 into training set (including devset) and test set. Some statistics are shown in~\ref{tab:datasp}. The training set contains 15216 positive news articles selected from 45 weeks (with weekly index difference $>$ 2\%) and 15778 negative news articles from 45 weeks (with weekly index difference $<$-2\%). To support multi-task learning that predicts the worthiness of news, we labelled 2000 positive and 2000 negative examples in terms of worthiness by either using category proxies or manually:

\begin{itemize}
\item Negative:~Basic-materials (250), Cyclicals (250), Non-cyclicals (250), and Healthcare (250), European (250) , Manually labelled (750);
\item Positive: Top 20 weighted companies news\footnote{\href{https://us.spindices.com/indices/equity/sp-500-top-50}{https://us.spindices.com/indices/equity/sp-500-top-50}} (750), Financials (500), U.S (250), Technology (250), Manually labelled (250).

\end{itemize}

The remaining data are used for constructing the data for the summarizer. For each week, 100 news articles are randomly selected and their sentiment scores are averaged as the overall sentiment score. The ground truth of each week's trend is labelled with different binning polices so that our model is comparable with other baselines. For example, for three category classification, we followed previous work~\citep{hu2018listening} and labelled our data as UP ($>$ 0.79\%), DOWN ($<$-0.21\%), or Preserve (-0.21\% $\sim$ 0.79\%) so that three classes are approximately even. Because the the weeks in which the news articles are used for training the sentiment extractor can longer be used for the summarizer, there are 407 weeks left for evaluation. We used 250 weeks for training and the rest for testing.  

\begin{table}[!htb]
\centering
\begin{adjustbox}{width=\columnwidth}
\begin{tabular}{cccc}
\hline
  & Positive ($>$2\%) & Negative ($<$-2\%) &  Neutral\\
\hline
\# weeks &  69 & 53 & 327 \\
\# news &  23319 & 18599 & 107352 \\
\# avg news/week &  338 & 351 & 328 \\
\hline 
\end{tabular}
\end{adjustbox}
\caption{Data labelling and distribution from August 2009 to April 2019. The 2\% threshold is only for selecting training examples for the sentiment extractor.}
\label{tab:datasp}
\end{table}

Following previous works~\citep{xie2013semantic,ding2015deep,xu2018stock}, we adopt two metrics including Accuracy (Acc) and Matthews Correlation Coefficient (MCC) scores to evaluate our model.

\subsection{Baselines}

For comparison, we select several public models
as baselines including: (1) \textbf{EB-CNN} \citep{ding2015deep}, a deep convolutional neural network model that predicts the index trend (binary, up or down) with short-term and long-term influences of events. (2) \textbf{StockNet} \citep{xu2018stock}, a hybrid RNN that takes both the text and past price information to predict the binary movement of the stock. In their approach, the neutral data are removed. (3) \textbf{HAN} \citep{hu2018listening}, a Hybrid Attention Networks to predict the
stock trend (Up, Down, and Preserve) based on the sequence of recent related news.
We also compare our model with its naive
versions in which the polarity-over-time representation or multi-task learning is not used. 

\subsection{Experimental Settings}

We adopt the BERT (base) model as the main architecture of the sentiment extractor and choose the hyper-parameters on the dev-set with grid search. For the polarity-of-time representation, we used the following set of parameters: \{V = 512, L = 4, $\alpha$ = 0.5\}. The representation of the CLS token (embedding size: 768) and the scaled polarity-over-time vector are concatenated and fed to a dense layer (512, activation='ReLU') then followed by two separate prediction heads (the sentiment prediction as well as the worthiness prediction, both use linear transformation with softmax). The weight of two losses is set to 0.5. The model is trained with a batch size of 32 and default learning rate of 2e-5. 

We choose the maximum input length after sub-word tokenizing to be 180. It is much smaller than the limit (512 tokens), however, such length can cover the headline and first few paragraphs - the most informative parts of the news.

\subsection{Effect of POT and Multitask Learning}

We use LSTM as the baseline for the sentiment extractor. As shown in Table~\ref{tab:se1}, BERT can better capture the nuance sentiment towards the market as compared to LSTM. The POT representation leads to significant improvements on our sentiment extractor. The word polarity that varies according to certain events provides global information and complements the BERT representation of news in which the text is truncated. The multi-task learning (MTL) module that additionally predicts the worthiness of the news, nonetheless, slightly improves the model performance by 1.4\%. This may result from the lack of labelled data and inaccurate labelling of data from using proxies. However, As both modules improves all
evaluation metrics, we use the BERT+POT+MTL as the default sentiment extractor in the following experiments.

\begin{table}[!htb]
\centering
\small
\begin{tabular}{lll}
\hline
\textbf{Models} & \textbf{Accuracy}  & \textbf{F1}\\
\hline
LSTM &  0.573 & 0.709\\
BERT &  0.635 & 0.726\\
BERT+POT &  0.693 & 0.754\\
BERT+POT+MTL &  0.707 & 0.761\\
\hline 
\end{tabular}

\caption{The performance of sentiment extractor with different modules.}
\label{tab:se1}
\end{table}

\begin{figure}[!htb]
    \centering 
    \includegraphics[width=\columnwidth]{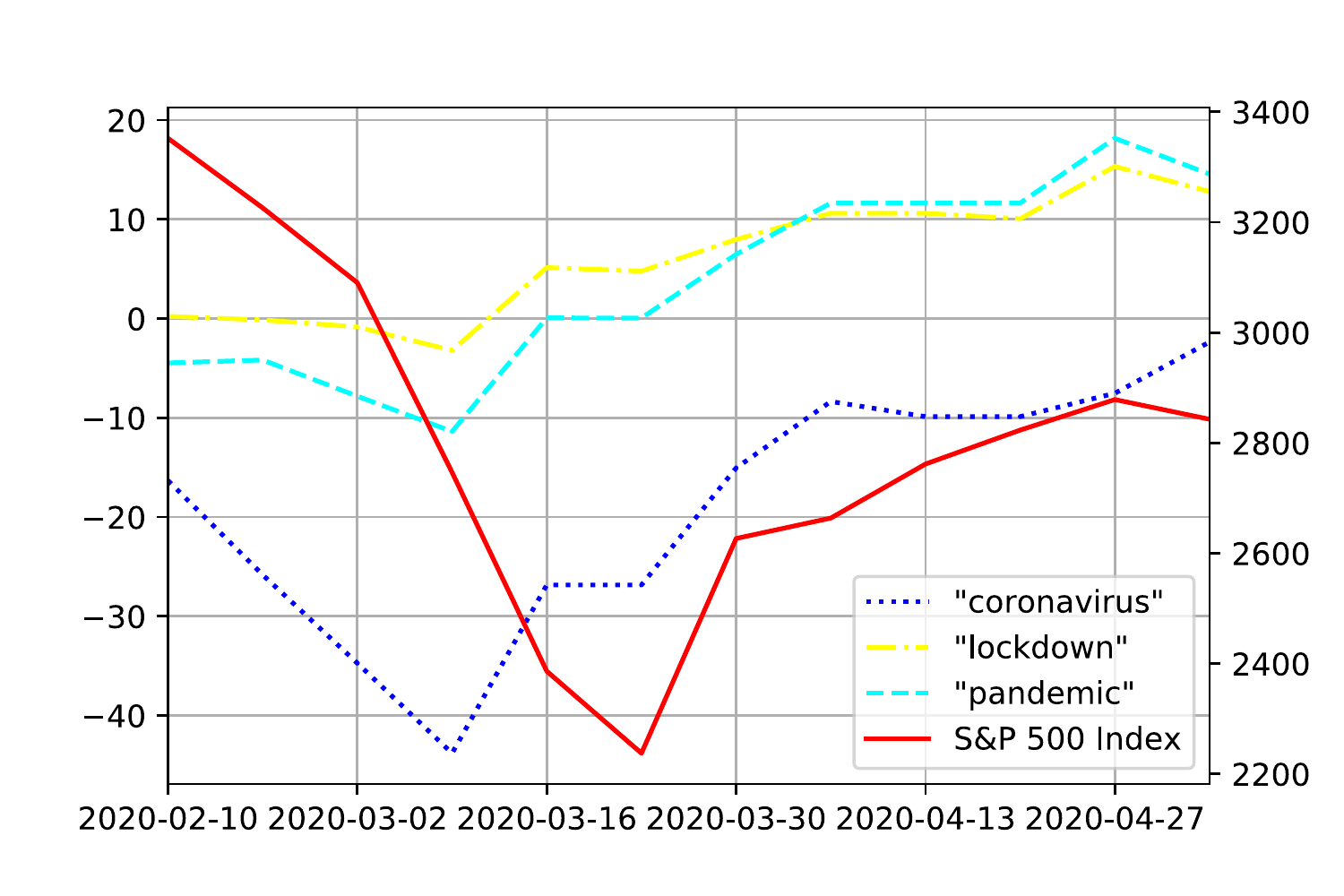}
    \caption{The change of the polarity-over-time (POT) score of ``coronavirus", ``lockdown", and ``pandemic" early in 2020. }
    \label{fig:pota}
\end{figure}

Further analysis is performed on POT to illustrate how the sentiment of certain words change with events and thus the stock index. As illustrated in Figure~\ref{fig:pota}, the term ``coronavirus" shows a strong negative signal to the stock market. However, the sentiments of ``lockdown" and ``pandemic" turning slightly positive indicate that the market is back to calm and the confidence is rebuilt from the acknowledgement to and the countermeasures against the epidemic. The term ``distancing", ``ventilator", and even ``covid" (such abbreviation also indicates the acceptance of the fact) show similar positive trend as ``lockdown" and ``pandemic". 

\begin{figure}[!htb]
    \centering 
    \includegraphics[width=\columnwidth]{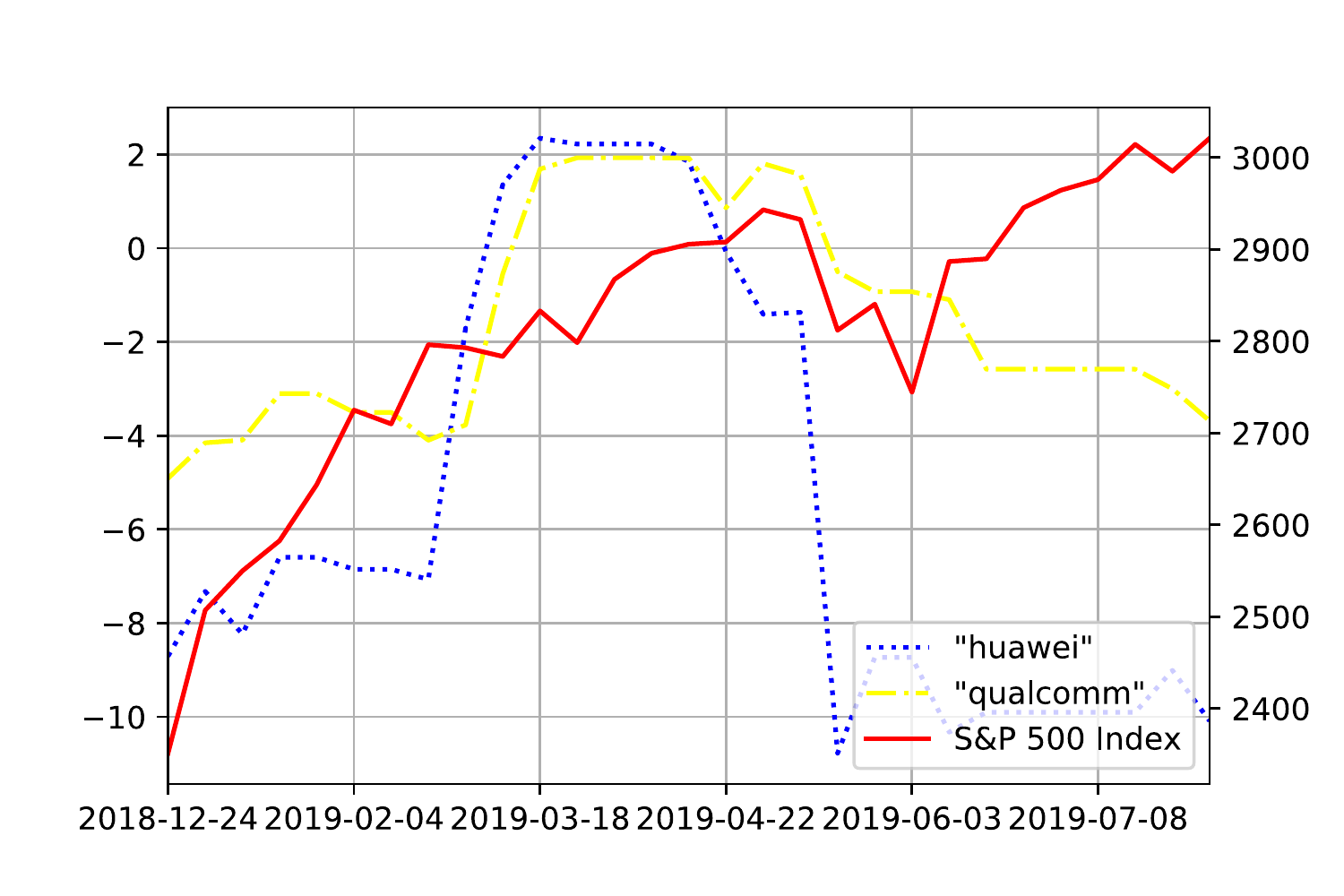}
    \caption{The event-driven fluctuations of the S\&P 500 Index and the POT scores of ``huawei" and ``qualcomm".}
    \label{fig:potc}
\end{figure}

Figure~\ref{fig:potc} shows the market uncertainty under the influence of the Government policy and the rival between Huawei and Qualcomm. Ban policy against Huawei during Feb 2019 and the extradition hearing for Huawei's CFO on March 2019 had a great negative impact on Huawei. However, this is a positive signal to the Qualcomm and the U.S. market. Therefore, the polarity of ``huawei" was positive during that period. Nevertheless, since May 2019, Huawei's countermeasures and other positive events launched a negative signal to Qualcomm and U.S. market\footnote{More detailed event list can be found \href{https://www.cnet.com/news/huawei-ban-full-timeline-us-restrictions-china-trump-executive-order-uk-ban-nokia-ericsson/}{here}.}. These event-driven impacts are well captured by the POT metric. 

\begin{figure}[!htb]
    \centering 
    \includegraphics[width=\columnwidth]{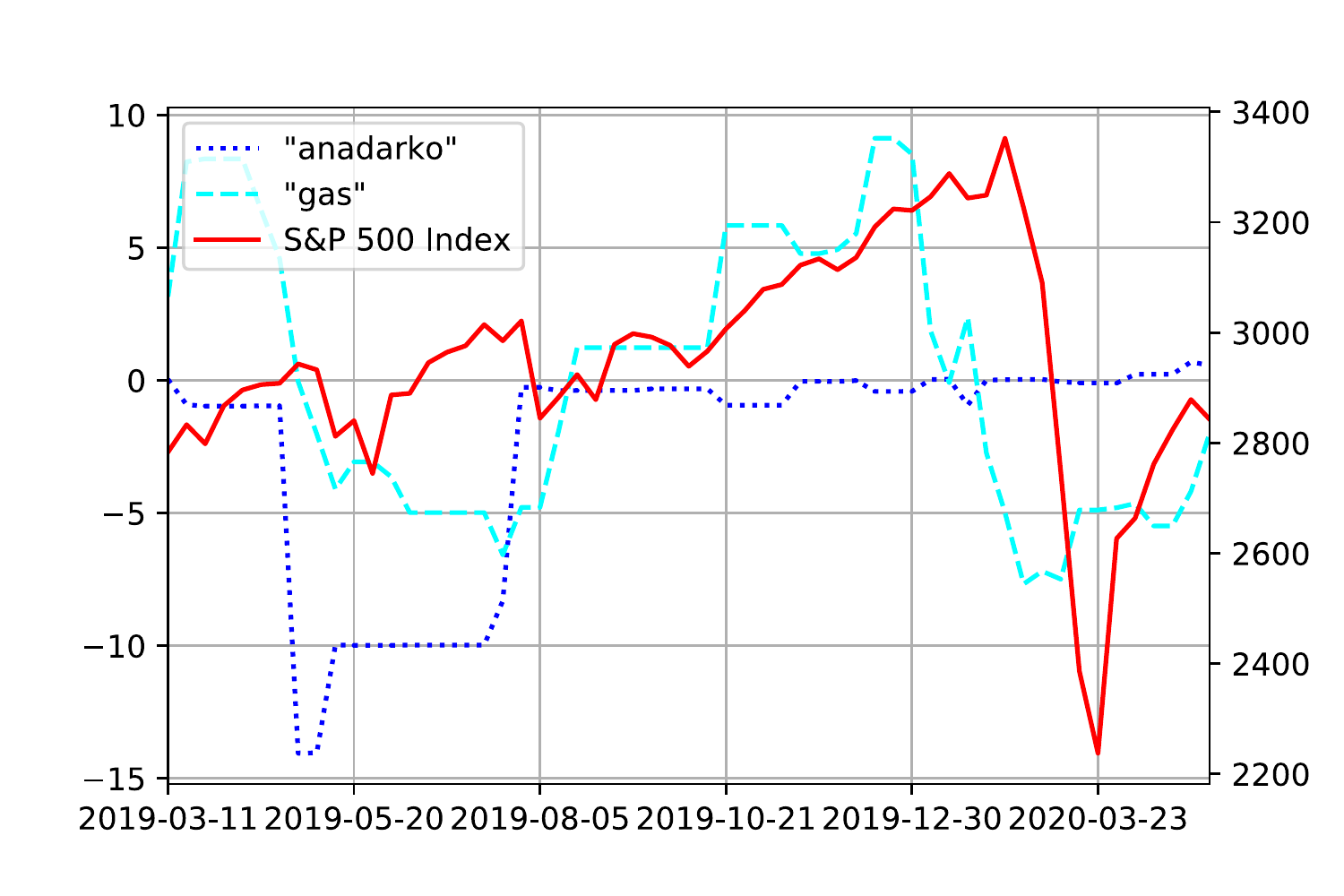}
    \caption{The POT scores show the negative events of Anadarko impacted the gas and stock market.}
    \label{fig:potb}
\end{figure}

The hit of the negative event of Anadarko Petroleum on the U.S. market is illustrated in Figure~\ref{fig:potb}. The event that Occidental was going hostile on Anadarko on April 2019 introduced a negative surprise to the market. Meanwhile, the polarity of ``gas" and the stock index experienced an unexpected drop. However, the polarity of ``gas" continues to vary according to other events, but the sentiment of ``anadarko" remains neutral since no big event released afterwards. 

\begin{table*}[!htb]
\centering
\begin{adjustbox}{width=\textwidth}
\begin{tabular}{lllll}
\hline
\textbf{Models} & \textbf{Accuracy}  & \textbf{MCC} & \textbf{Types} & \textbf{Binning Policies} \\
\hline 
EB-CNN~\citep{ding2015deep} & 0.6421 & 0.4 & Binary  & Up: p\% $>$ 0; Down: p\% $<$ 0 \\
StockNet~\citep{xu2018stock}  & 0.5823 & 0.08 & Binary & Up: p\% $>$ 0.55; Down: p\% $<$ -0.5 \\
HAN~\citep{hu2018listening}  & 0.478 & - & 3 Categories & Up: p\% $>$ 0.87; Down: p\% $<$ -0.41;  Otherwise: Preserve \\ 
\hline 
OUR+Naive & 0.4824 & 0.2573 & \multirow{2}*{3 Categories }  & \multirow{2}*{Up: p\% $>$ 0.79; Down: p\% $<$ -0.21;  Otherwise: Preserve}  \\
OUR+Full & 0.4976 & 0.2795 &  &  \\ 
\hline 
OUR+Full & 0.6883 & 0.3767 & \multirow{2}*{Binary}   & Up: p\% $>$ 0; Down: p\% $<$ 0 (Asymmetrically binned) \\
OUR+Full & 0.6526 & 0.3101 &    & Up: p\% $>$ 0.6; Down: p\% $<$ 0 (Symmetrically binned) \\
\hline 
\end{tabular}
\end{adjustbox}
\caption{Experimental results and comparisons with other baselines. The OUR+Naive refers to the model that only uses BERT, while the OUR+Full model utilizes discrete word features and multi-task learning. It can be difficult to compare our model with tweet-based StockNet since they do not provide the leader board with the news corpus.}
\label{tab:full1}
\end{table*}

\subsection{Overall Results}

In this part, we will show the performance of our full model and provide
some further analysis.  From the results shown in Table~\ref{tab:full1}, we can observe that: 

(1) Our full model using discrete word polarity feature and multitask learning achieves better performance on predicting the weekly index trend than the naive version of our model which only uses BERT. The word polarity over time and the worthiness of a news article provide meaningful information to both the sentiment extractor and summarizer. 

(2) We first run our task as multiclass classification with three categories balanced as~\citet{hu2018listening}. Both of our naive and full model show considerable improvement as compared to their approach. It not only indicates the effectiveness of our model architecture but also justifies the weekly-Monday framework for sentiment-based stock prediction. Our full model also shows a significant improvement as binary classification compared with EB-CNN.

(3) Although, our model works well with the weekly-monday basis, its performance is worsen with daily prediction. The highly volatile daily price variations and lag of the market reacts on sentiment signals restrain the model performance. 

\begin{figure}[!htb]
    \centering 
    \includegraphics[width=\columnwidth]{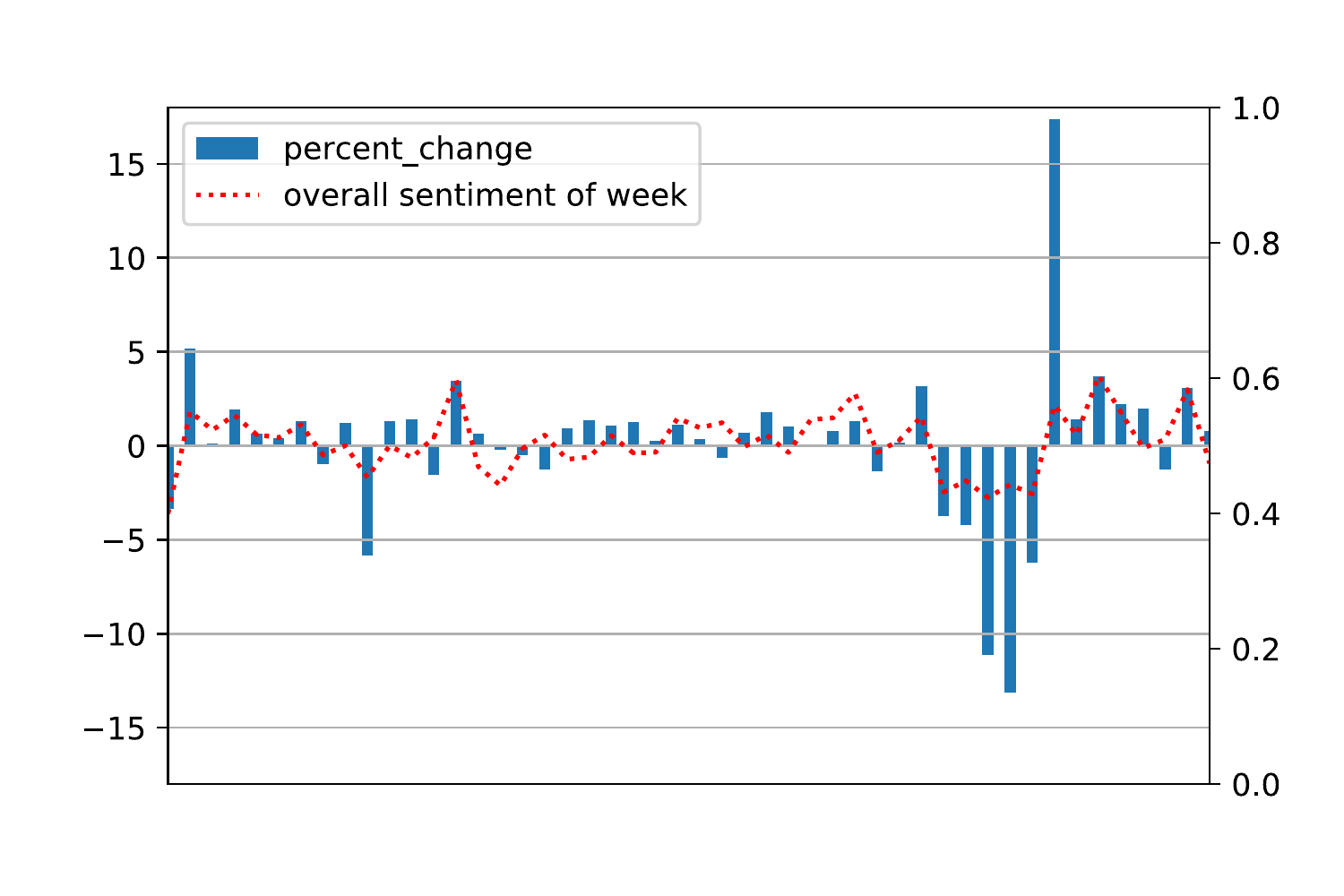}
    \caption{Index percent changes and overall sentiment scores of different weeks from May 2019 to May 2020. The graphic is composed from an overlay of the S\&P 500 weekly percent changes and the overall sentiment scores from the week prior.}
    \label{fig:sum1}
\end{figure}

In addition, Figure \ref{fig:sum1} illustrate how the predicted overall sentiments of different weeks relate to the weekly index changes. We select a one-year period from May 2019 to May 2020 containing minor and major weekly index changes. The overall sentiment scores summarized from the individual news sentiments range from 0 to 1, and the weekly index changes over the past 10 years are between -13\% and 17\%. The sentiment score of a certain week and index changes of $next$ week are highly correlated with a correlation coefficient of 0.65. On average, the higher the sentiment score, the higher the probability that the index will increase next week. However, such correlation suffers from the weeks with minor index changes.

\section{Conclusion and Future Work}

In this paper, we propose a sentiment-based stock index prediction system
which contains a sentiment extractor that distills the polarity of the news articles towards the market and a summarizer that sums up the overall sentiment of the week to predict the index change of next week. We propose a discrete word feature called Polarity-Over-Time (POT) which captures the sentiment changes of words according to certain events at different periods of time. Both the POT feature and multi-task learning are used to improve the performance of the sentiment extractor. We show that our
model on the 10-year Reuters news dataset achieves considerable improvements as compared to other baselines. In particular, we
demonstrate that the weekly-Monday framework provides space for the market to react on sentiment signals and therefore, is appropriate for sentiment-based stock prediction.

In the future, we will explore the following directions:
(1) The adaptation of our model to daily price prediction. We will explore how to adapt our model to predict the daily price variations in which stronger denoising methods may be applied to sentiment extraction.

(2) The merge of the sentiment extractor with the summarizer as one integrated neutral architecture such as BERT with hierarchical CNN so that the parameters can be jointly learned and layers instead of probability scores can be fed to the summarizer.

\bibliography{anthology,acl2020}
\bibliographystyle{acl_natbib}

\end{document}